\documentclass[runningheads]{llncs}
\usepackage[T1]{fontenc}
\usepackage{graphicx}
\usepackage{algorithm}  
\usepackage{algpseudocode}  
\usepackage{amsmath} 
\usepackage[colorlinks, linkcolor=red, anchorcolor=blue, citecolor=blue]{hyperref}

\begin{document}
\title{ Spatio-temporal Graph Learning on Adaptive Mined Key Frames for High-performance Multi-Object Tracking 
\thanks{Corresponding author: Xiao Wang (\email{xiaowang@ahu.edu.cn})} 
}

\author{
Futian Wang \inst{1,2} \and 
Fengxiang Liu \inst{1} \and 
Xiao Wang* \inst{1,2} 
}      

\authorrunning{Fengxiang Liu et al. }

\institute{
1. School of Computer Science and Technology, Anhui University, Hefei, China \\ 
2. Anhui Provincial Key Laboratory of Multimodal Cognitive Computation, Hefei, China 
}

%
%

\maketitle              
\begin{abstract}

In the realm of multi-object tracking, the challenge of accurately capturing the spatial and temporal relationships between objects in video sequences remains a significant hurdle. This is further complicated by frequent occurrences of mutual occlusions among objects, which can lead to tracking errors and reduced performance in existing methods. Motivated by these challenges, we propose a novel adaptive key frame mining strategy that addresses the limitations of current tracking approaches. 
Specifically, we introduce a Key Frame Extraction (KFE) module that leverages reinforcement learning to adaptively segment videos, thereby guiding the tracker to exploit the intrinsic logic of the video content. This approach allows us to capture structured spatial relationships between different objects as well as the temporal relationships of objects across frames. 
To tackle the issue of object occlusions, we have developed an Intra-Frame Feature Fusion (IFF) module. Unlike traditional graph-based methods that primarily focus on inter-frame feature fusion, our IFF module uses a Graph Convolutional Network (GCN) to facilitate information exchange between the target and surrounding objects within a frame. This innovation significantly enhances target distinguishability and mitigates tracking loss and appearance similarity due to occlusions. 
By combining the strengths of both long and short trajectories and considering the spatial relationships between objects, our proposed tracker achieves impressive results on the MOT17 dataset, i.e., 68.6 HOTA, 81.0 IDF1, 66.6 AssA, and 893 IDS, proving its effectiveness and accuracy. 

\end{abstract}

\section{Introduction}

\begin{figure}
    \centering
    \includegraphics[width=0.85\linewidth]{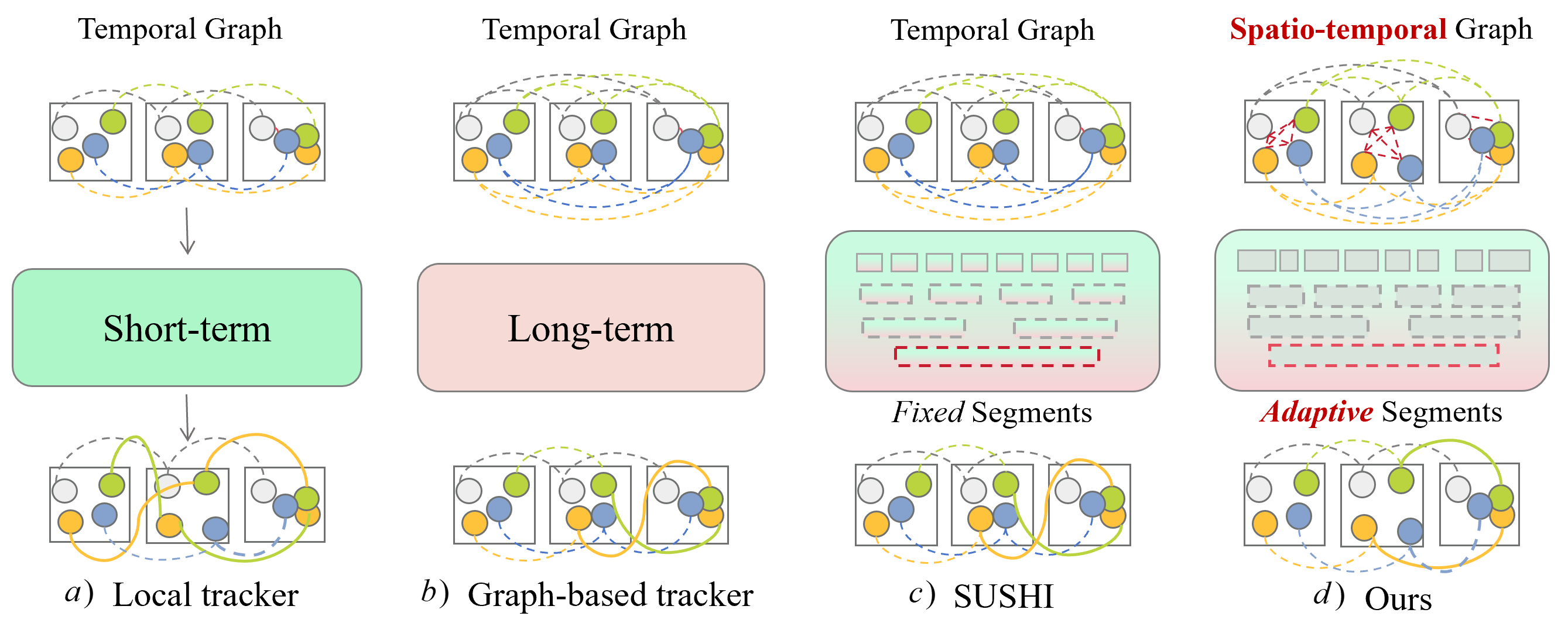}
    \caption{Comparison between (a, b, c) existing algorithms and (d) our newly proposed MOT tracking framework.}
    \label{fig:firstIMG}
\end{figure}

Multi-Object Tracking (MOT) aims at tracking multiple objects simultaneously in a video, maintaining the identity of the objects, and generating their motion trajectories. It has wide applications in different fields such as video surveillance, autonomous driving, and video analytics. Despite many approaches proposed for multi-object tracking, fragmented trajectories or ID Switching (IDS) problems caused by frequent occlusions in crowded scenes and similar appearances are still significant challenges.

\begin{figure}
\centering
\includegraphics[width=0.85\linewidth]{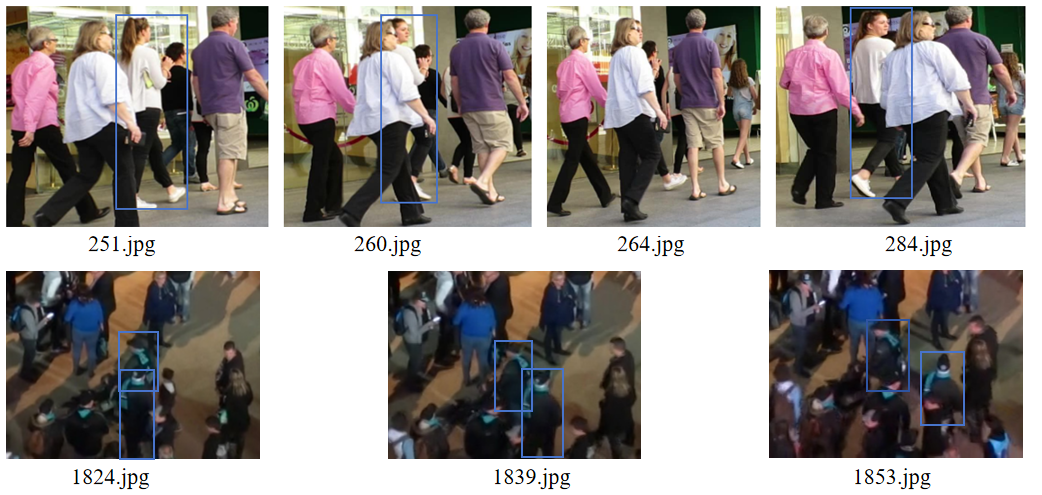}
\caption{In the top row of the image, the individuals represented are extracted from the objects within the MOT17-09 video sequence of the MOT17 dataset. It is worth noting that the girl with a high ponytail reappears multiple times, each time experiencing varying degrees of occlusion. In the second row, the two highlighted objects in the image exhibit a high degree of visual similarity and are located in close proximity to one another. To distinguish between these two objects, we propose utilizing contextual information from the surrounding objects.} 
\label{fig:video_seq_fig}
\end{figure}

To address the ID Switching (IDS) issue, most state-of-the-art trackers use a combination of short and long trajectories for tracking~\cite{gao2024multi}~\cite{cetintas2023unifying}. Due to the variability of scenarios, different specific schemes are required, for example, motion-based local trackers and appearance-based trackers~\cite{xiao2024motiontrack}, \cite{yi2024ucmctrack}, \cite{zhang2021fairmot}, \cite{zhao2023representation}. When there is heavy occlusion and uniformity of clothing, these methods often become highly specific to particular scenarios, making them not easily scalable to broader applications. Gao et al.~\cite{gao2024multi} propose a hierarchical approach to processing video, with lower levels focusing on short-term associations and higher levels focusing on increasingly long-term scenes. 
Cetintas et al.~\cite{cetintas2023unifying} uses the same learnable model for all time scales, thus improving scalability. Many state-of-the-art MOT trackers use graph neural networks~\cite{kipf2016semi} to handle similar appearance problems~\cite{cheng2023rest}~\cite{you2023utm}. However, there are still some issues that exist in current works, e.g., many of these methods are highly specific. Cetintas et al.~\cite{cetintas2023unifying} use the same learnable model for all time scales, obviously, the scalability and flexibility can be improved. In addition, we are also inspired by the scenarios as shown in Fig.~\ref{fig:video_seq_fig}. More in detail, scenario 1 demonstrates that the time interval between the appearance of a tagged object and its occlusion to reappearance is not consistent. There is a tendency in scenario 2 to follow the wrong phenomenon due to similar clothing and similar location. However, few existing MOT systems take into account the spatial relationships between different objects within a single frame.

\begin{figure}
\centering
\includegraphics[width=0.85\linewidth]{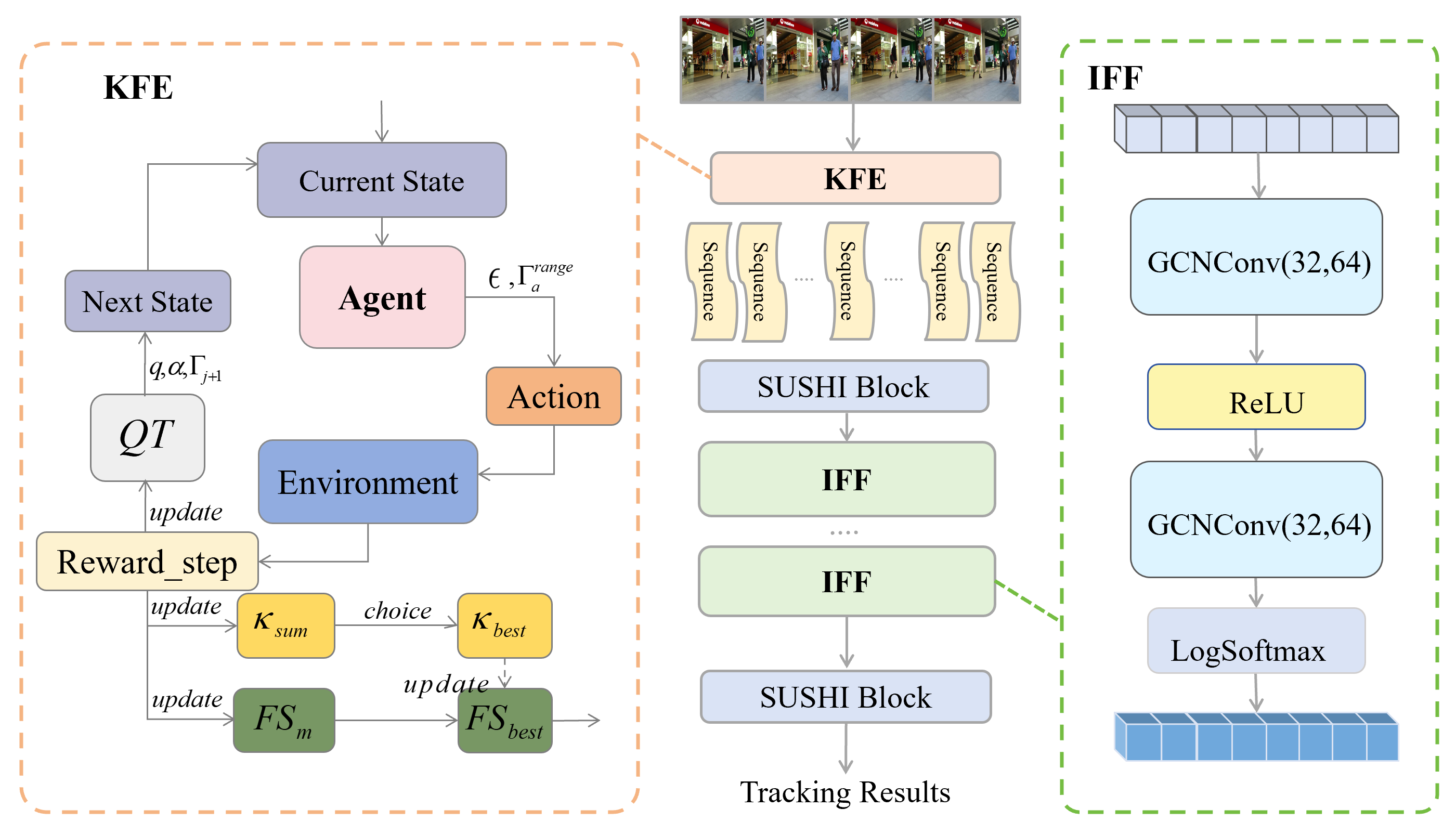}
\caption{An overview of our proposed multi-object tracking framework.} 
\label{fig:framework} 
\end{figure}

To address the issues mentioned above, in this paper, we propose a novel multi-object tracking by spatio-temporal graph learning on adaptive mined key frames. As shown in Fig.~\ref{fig:framework}, given the input video frames, we first adopt a KFE (Key Frame Extraction) module to divide them into frame segments adaptively. The key frame extraction is conceptualized as a decision-making process through the application of the Q-learning algorithm, which capitalizes on the high performance capabilities of short trajectories. Furthermore, we harness spatio-temporal graph learning in conjunction with the Intra-frame Feature Fusion (IFF) module to amplify the interactions among targets within the same frame and adjacent frames. By integrating these two modules into the baseline tracker SUSHI~\cite{cetintas2023unifying}, the overall tracking performance is significantly augmented.

To sum up, we draw the main contributions of this paper as the following three aspects:
\begin{itemize}
\item We propose a novel adaptive key frame mining strategy guided multi-object tracking algorithm based on reinforcement learning. It is a unified, scalable, and hierarchical tracker that models both short-term and long-term associations simultaneously. 

\item We propose a new spatio-temporal graph learning module that captures the structured spatial relations between different pedestrians in a single frame and temporal relations between different frames. 

\item Extensive experiments on the public MOT17 benchmark dataset fully validated the effectiveness of our proposed strategies for the MOT task. 
\end{itemize}



\section{Methodology}


\subsection{Reinforcement Learning based Key Frame Mining}
The KFE module is based on Q-learning, and we need to design the corresponding action selection, reward mechanism, active exploration strength, and table update strategy. The optimal segmentation strategy $FS_{best}$ and the corresponding optimal reward score $\kappa_{best}$ are iterated during the learning process. The action selection is divided into the first frame of the video segment $\Gamma_{a}$, and the last frame action selection $\Gamma_{b}$, both of which have the same selection strategy, in order to avoid redundancy, we only illustrate the action selection strategy for the first frame. $\Gamma^{i+1}$ is the action selection used to record the choice that will be made next, $QT$ is the Q-value used to record the total return expected to be obtained by taking a specific action in a certain state, $F$ denotes the state, $QT[F_{a}^{i}, \Gamma_{a}^{j}]$ represents the expected return of the next choice $\Gamma_{a}^{j+1}$ in state $F_{a}^{i}$, and  $\Gamma_{a}^{range}$ represents the ranges of values of $\Gamma_{a}$. we choose the action with the highest reported value in the Q-value as the next action, in addition to randomly choosing the next action with a certain exploration rate $\epsilon$.
\begin{align}
\Gamma_{a}^{j+1} =   
\left\{  
\begin{array}{ll}  
\max(QT[F_{a}^{i}, \Gamma_{a}^{j}]), &\eta < \epsilon \\  
random(\Gamma_{a}^{range}), & \text{otherwise}  
\end{array}  
\right. 
\end{align}
where $\eta\in(0, 1), j\in(0, N)$ and $\epsilon=0.1$.
We need to develop a reasonable reward mechanism to determine the direction of model optimization, we use the feature difference between the first and last frames between the video segments as the reward value, $\kappa_{i+1}$ is the reward for the $i+1$th segmentation result of the current round. 
\begin{align}
\kappa_{i+1} =((1-\varphi(f_{F_{a}^{i}}, f_{F_{a}^{i+1}})+(1-\varphi(f_{F_{b}^{i}}, f_{F_{b}^{i+1}}))\times\delta+\xi   
\end{align}
where $f_{F_{a}^{i}}$ represents the feature of the first frame of the $i$th video segment, $f_{F_{b}^{i}}$ represents the feature of the last frame of the $i$th segment, $f_{F_{a}^{j}}$ represents the feature of the first frame of the $j$th video segment, and $f_{F_{b}^{j}}$ represents the feature of the last frame of the $j$th segment. Additionally, $\delta$, $\xi$ are constants, and $\varphi$ denotes the cosine similarity. 
Regarding the update of the Q-value, we follow the approach of classical algorithms, where the Q-value is dependent on $F^{i}$, $F^{i+1}$, the learning rate $q=0.1$, the reward $\Gamma$, and the discount factor $\alpha=0.99$.
\begin{align}
\Gamma_{best}^{i+1} =max(QT(F^{i+1}, \Gamma^{i})
\end{align}
\begin{align}
QT[F^{i}, \Gamma]=\lambda\times q+QT[F^{i}, \Gamma]
\end{align}
where $\Gamma_{best}^{i+1}$ represents the optimal action selection for the next step, $\lambda=\kappa_{i}+\alpha\times QT[F^{i+1}, \Gamma_{best}^{i+1}]-QT[F^{i}, \Gamma]$.

When the video is fully segmented after the $m$th iteration, the final reward score for this round $\kappa_{sum}$ is denoted as
\begin{align}
    \kappa_{sum}=\frac{\sum_{i=1}^{n} \kappa_{a}^{i}}{len(FS_{m})}
\end{align}
where $\kappa_{a}^{i}$ is the reward for the $i$th segmentation result of the first frame of the current round and $FS_{m}$ is the segmentation strategy in the $m$th iteration. We evaluate whether the current segmentation result is attributed to the advanced optimal reward $\kappa_{best}$. If it is superior, we proceed to update the parameters $FS_{best}$ and $\kappa_{best}$ accordingly
\begin{equation}  
\left\{  
\begin{array}{ll}  
\kappa_{best}=max(\kappa_{best}, \kappa_{sum}) \\  
FS_{best}=max(FS_{best}, FS_{m})\\  
\end{array}  
\right. 
\end{equation}

Through the aforementioned steps, we are able to achieve efficient segmentation results that fully exploit the inherent logic of the video, effectively combining the advantages of both short and long trajectories. Compared to SUSHI, our approach achieves up to +2.5 AssA and up to +1.7 IDF1 improvements, with a decrease in IDS. This indicates that the KFE module can effectively mitigate the issue of identity loss when objects reappear after occlusion. Thereby validating the effectiveness of our method.The complete inference procedure is outlined in Algorithm 1, where $\chi_{a}^{j}$ represents the state selection of the first frame in the $j$th video, $\chi_{b}^{j}$ represents the state selection of the last frame in the $j$th video.

\begin{algorithm}  
\caption{KFE Algorithm}  
\small 
\begin{algorithmic}[1]  
\Require 
    \Statex env: The environment with two state spaces and action spaces  
    \Statex agent: A QAgent instance  
    \Statex episodes: The number of episodes to train  
    \Statex $\epsilon$: The exploration rate  
    \Statex $q$: The learning rate for Q-table updates  
    \Statex $\alpha$: The discount factor for future rewards  
    \Statex $\Gamma_{a}^{range}$: The range for selecting future state  
\Ensure 
    \Statex $FS_{best}$: The states recorded during the best episode  
  
\State Initialize $\kappa_{best}$,best\_episode, $FS_{best}$  
\For{episode = 1 to episodes}  
    \State Initialize $\chi_{a}^{j}$,$\chi_{b}^{j}$,done, $\kappa_{sum}$,$FS_{m}$  
    \While{not done}  
        \State $\Gamma_{a}^{j+1} \gets$ agent. select\_action($\chi_{a}^{j}$,$\epsilon$)  
        \State $\Gamma_{b}^{j+1} \gets$ agent. select\_endaction($\chi_{b}^{j}$,$\epsilon$)  
        \State $(\chi_{a}^{j+1},\chi_{b}^{j+1}$,$\kappa_{a}^{j+1}$,$\kappa_{b}^{j+1}$,done) $\gets$ env. step($\Gamma_{a}^{j+1}$,$\Gamma_{b}^{j+1}$)  
        \State agent. update\_q\_table($\chi_{a}^{j}$,$\Gamma_{a}^{j+1}$,$\kappa_{a}^{j+1}$,$\chi_{a}^{j+1}$,$q$,$\alpha$)  
        \State agent. update\_qend\_table($\chi_{b}^{j}$,$\Gamma_{b}^{j+1}$,$\kappa_{b}^{j+1}$,$\chi_{b}^{j+1}$,$q$,$\alpha$)  
        \State $\chi_{a}^{j} \gets \chi_{a}^{j+1}$  
        \State $\chi_{b}^{j} \gets \chi_{b}^{j+1}$  
        \State $\kappa_{sum} \gets \kappa_{sum} + \kappa_{a}^{j+1} + \kappa_{b}^{j+1}$  
        \If{not done}  
            \State $FS_{m}$.append(($\chi_{a}^{j+1}$,$\chi_{b}^{j+1}$))  
        \EndIf  
    \EndWhile  
    \If{$\kappa_{sum} > \kappa_{best}$}  
        \State $\kappa_{best} \gets \kappa_{sum}$  
        \State best\_episode $\gets$ episode  
        \State $FS_{best} \gets FS_{m}$  
    \EndIf  
\EndFor  
\State \Return $FS_{best}$  
\end{algorithmic}  
\end{algorithm}

\subsection{Spatial-temporal Relation Mining} 
When occlusion of an object occurs, we can combine the surrounding objects to perform object recognition where the objects are too similar to each other and their positions are close. Inspired by the above ideas, we add IFF to solve the problem of occlusion and similar appearance. In performing the fusion of different levels, we first construct a graph $G=(V, E)$ for the level, where the object is represented by node $v_{i}\in V$ and the edge $e_{i}$ represent hypotheses for judging the relationship between nodes. We perform in-frame feature fusion before hierarchy merging. The initial scheme is to complement the features of the object with a certain percentage after averaging the surrounding objects.
    \begin{align}
        f_{v_{i}}=a \times f_{v_{i}} + b \times avg(f_{V_{i}^{m}})
    \end{align}
where $f_{v_{i}}$ represents the feature of  node $v_{i}$, $V_{i}^{m}$ represents the set of  nodes closest to $v_{i}, a, b\in (0, 1), b=1-a$.
However, GCN is able to capture more complex contextual information than the simple average through the information transfer and aggregation between nodes, adaptively learns the relationship weights between different objects, and can automatically adjust the degree of influence of different neighboring nodes on the features of the central node according to the actual situation, so as to more accurately reflect the real relationship between the objects. In the case where the object is occluded, GCN can supplement the feature representation of the occluded object by fusing the information of the surrounding unoccluded objects, which helps to alleviate the impact of occlusion on recognition performance. When the objects are too similar to each other and close to each other in terms of location, GCN can utilize the structured spatial relations between objects. Therefore, we chose the method GCN for feature fusion with graphs.
    \begin{align}
        f_{v_{i}}=a \times f_{v_{i}} + b \times GCN(f_{v_{i}}, f_{V_{i}^{m}})
    \end{align}
Utilizing GCN, we integrate features from neighboring objects, enabling each object to encapsulate a more comprehensive global context. This approach leads to improvements of up to +2.0 in AssA and +1.1 in IDF1, empirically validating the efficacy of our method within the realm of computer vision research.

\subsection{Loss Function}
Our proposed KFE  method stands independent of the SUSHI Block and IFF frameworks. To determine the number of learning epochs, denoted as $M=(LN-u)\times(LN-n)\times100$,  where $u$ represents the shortest length of the video segment, $n$ represents the maximum length of the video segment, and $LN$ represents the total length of the video. We adopt the unfreezing strategy from the SUSHI Block, where subsequent levels are unfrozen after 500 iterations. We apply the focal loss on the generated edge classification scores and sum these losses across all levels to obtain the final loss~\cite{lin2017focal}. 

\section{Experiments}

\subsection{Datasets and Evaluation Metrics} 

We conducted our experiments on the public dataset MOT17. 
We used MOT17-Private for experimental effect validation. We follow the HOTA protocol~\cite{luiten2021hota} for quantitative evaluation, where HOTA focuses on overall tracking quality, AssA is used to measure association accuracy, and IDS focuses on measuring the number of identity switches during tracking. In addition, we also adopt the metrics MOTA and IDF1, which reflect the overall performance of detection and tracking as well as association accuracy, respectively, providing us with a multidimensional evaluation reference.


\subsection{Ablation Study}
To demonstrate the effectiveness of the proposed modules, in Table~\ref{tab1} we conducted experiments on MOT17, where we describe the added features as: KFE (Key Frame Extraction), and IFF (Intra-frame Feature Fusion). We found that KFE and IFE improved the performance on the MOT17 set, where the IFF module improved by 1.1 HOTA and KFN improved by 1.2 HOTA. Adding both modules together improves 1.6 HOTA, 2.3 IDF1, and 3.3 AssA. 
\begin{table}
\centering
\caption{Validating Module Effectiveness. }\label{tab1}
\begin{tabular}{llllll}
\hline
 & HOTA↑ &  IDF1↑ & AssA↑ &MOTA↑ & IDS↓ \\
\hline
SUSHI	&67.0	&78.7	&62.9	&\textbf{85.1}	&930\\
SUSHI+IFF	&68.1	&79.8	&64.9	&84.8	&915\\
SUSHI+DP	&67.8	&79.7	&64.8	&84.8	&924\\
SUSHI+KFE	&68.2	&80.4	&65.4	&85.0	&855\\
\textbf{Ours} &\textbf{68.6}	&\textbf{81.0}	&\textbf{66.6}	&84.0	&\textbf{893}\\
\hline
\end{tabular}
\end{table}

In Tab.~\ref{tab1} we started to use dynamic programming (DP) for segmentation, which has poor learnability and limited scalability, so we used the reinforcement learning Q-learning method, which has good applicability and scalability, and can be obtained from the experiment, Q-learning is more effective and improves the performance of the algorithm by 1.2 HOTA, 1.7 IDF1, and IDS decreased, so we finally chose the Q-learning algorithm. In the IFF module, there is a fusion between the Intra-frame features, we used a fixed ratio, and after the experiment in Tab.~\ref{tab2}, we finally set a=0.4.

\begin{table} 
\caption{Ablation experiments on feature fusion with varying proportions. }\label{tab2}
\centering
\begin{tabular}{llllll}
\hline
 (a, b)& HOTA↑ &  IDF1↑ & AssA↑ &MOTA↑ & IDS↓ \\
\hline
(0, 1.0)	&67.0	&78.7	&62.9	&85.1	&930\\
(0.1, 0.9)	&67.2	&78.0	&63.7	&84.9	&963\\
(0.2, 0.8)	&67.6	&79.2	&63.9	&84.7	&921 \\
(0.3, 0.7)	&68.0	&\textbf{80.3}	&64.2	&85.0	&936\\  
\textbf{(0.4, 0.6)}	&\textbf{68.1}	&79.8	&\textbf{64.9}	&84.8	&915\\
(0.5, 0.5)	&67.8	&78.9	&63.9	&84.7	&987\\  
(0.6, 0.4)	&67.9	&79.7	&64.3	&85.0	&939\\  
(0.7, 0.3)	&67.8	&79.2	&63.6	&\textbf{85.2}	&939 \\ 
(0.8, 0.2)	&67.4	&78.9	&63.4	&\textbf{85.2}	&912\\
(0.9, 0.1)	&67.2	&78.7	&63.2	&\textbf{85.2}	&\textbf{909}\\
(1.0, 0)	&67.0	&78.6	&63.5	&84.4	&\textbf{909}\\
\hline
\end{tabular}
\end{table}

\subsection{Comparison on Public Benchmarks}
We conducted experiments on the MOT17 \cite{dendorfer2021motchallenge} public dataset. As shown in Tab.~\ref{tab3} our tracker achieves 68.6 HOTA, and our method outperforms all methods under HOTA ordering in the MOT- challenge. On the dataset, using the same detection, our method outperforms existing comparisons including SUSHI~\cite{cetintas2023unifying}, CoNo-Link~\cite{gao2024multi}, ByteTrack~\cite{zhang2022bytetrack}, OC-SORT~\cite{cao2023observation} and StrongSORT++~\cite{du2023strongsort}, where StrongSORT++ is an offline version of StrongSORT that enhances trajectories by post-processing tracking offline. The benchmark results strongly demonstrate the advanced performance of our tracker. 
\begin{table}
\caption{Test set results on MOT17 benchmark. Our tracker achieves State-Of-The-Art (SOTA) performance in the HOTA metric. }\label{tab3}
\centering
\begin{tabular}{llllll}
\hline
 & HOTA↑ &  IDF1↑ & AssA↑ &MOTA↑ & IDS↓ \\
\hline
ByteTrack~\cite{zhang2022bytetrack}	&63.1	&77.3	&62.0	&80.3	&2196\\
FairMOT~\cite{zhang2021fairmot}	&59.3	&72.3	&58.0	&73.7	&3303 \\
OC-SORT~\cite{cao2023observation}	&63.2	&77.5	&63.2	&78.0	&1950 \\
*StrongSORT++~\cite{du2023strongsort}	&64.4	&79.5	&64.4	&79.6	&1194 \\
CoNo-Link~\cite{gao2024multi}	&67.6	&79.3	&65.6	&\textbf{86.5}	&909 \\
SUSHI~\cite{cetintas2023unifying}	&67.0	&78.7	&62.9	&85.1	&930 \\
\textbf{Ours} &\textbf{68.6}	&\textbf{81.0}	&\textbf{66.6}	&84.0	&\textbf{893} \\
\hline
\end{tabular}
\end{table}


\section{Conclusion}
In this study, we start with the offline multi-object tracking algorithm SUSHI and propose a new approach of adaptive video segmentation combined with Intra-frame surrounding object feature complementation. The proposed KFE takes full advantage of short trajectories and solves the IDS problem to some extent, and the feature complementation of the surrounding environment within the frame of the IFF module makes the object more informative, thus solving the problem of following the wrong object due to occlusion or similarity in appearance. Both modules improve the performance of multi-object tracking in terms of correlation in a good way. 


\noindent \textbf{Acknowledgment}: This work is supported by the University Synergy Innovation Program of Anhui Province under Grant(No. GXXT-2022-042), Anhui Province Higher Education Scientific Research Project (No.2023AH052574), National Natural Science Foundation of China under Grant 62102205, the Anhui Provincial Natural Science Foundation under Grant 2408085Y032. The authors acknowledge the High-performance Computing Platform of Anhui University for providing computing resources.

\nocite{*}
\bibliography{myReference}

\begin{thebibliography}{10}
\providecommand{\url}[1]{\texttt{#1}}
\providecommand{\urlprefix}{URL }
\providecommand{\doi}[1]{https://doi.org/#1}

\bibitem{cao2023observation}
Cao, J., Pang, J., Weng, X., Khirodkar, R., Kitani, K.: Observation-centric
  sort: Rethinking sort for robust multi-object tracking. In: Proceedings of
  the IEEE/CVF conference on computer vision and pattern recognition. pp.
  9686--9696 (2023)

\bibitem{cetintas2023unifying}
Cetintas, O., Bras{\'o}, G., Leal-Taix{\'e}, L.: Unifying short and long-term
  tracking with graph hierarchies. In: Proceedings of the IEEE/CVF Conference
  on Computer Vision and Pattern Recognition. pp. 22877--22887 (2023)

\bibitem{cheng2023rest}
Cheng, C.C., Qiu, M.X., Chiang, C.K., Lai, S.H.: Rest: A reconfigurable
  spatial-temporal graph model for multi-camera multi-object tracking. In:
  Proceedings of the IEEE/CVF International Conference on Computer Vision. pp.
  10051--10060 (2023)

\bibitem{dendorfer2021motchallenge}
Dendorfer, P., Osep, A., Milan, A., Schindler, K., Cremers, D., Reid, I., Roth,
  S., Leal-Taix{\'e}, L.: Motchallenge: A benchmark for single-camera multiple
  target tracking. International Journal of Computer Vision  \textbf{129},
  845--881 (2021)

\bibitem{du2023strongsort}
Du, Y., Zhao, Z., Song, Y., Zhao, Y., Su, F., Gong, T., Meng, H.: Strongsort:
  Make deepsort great again. IEEE Transactions on Multimedia  \textbf{25},
  8725--8737 (2023)

\bibitem{gao2024multi}
Gao, Y., Xu, H., Li, J., Wang, N., Gao, X.: Multi-scene generalized trajectory
  global graph solver with composite nodes for multiple object tracking. In:
  Proceedings of the AAAI Conference on Artificial Intelligence. vol.~38, pp.
  1842--1850 (2024)

\bibitem{kipf2016semi}
Kipf, T.N., Welling, M.: Semi-supervised classification with graph
  convolutional networks. arXiv preprint arXiv:1609.02907  (2016)

\bibitem{lin2017focal}
Lin, T.: Focal loss for dense object detection. arXiv preprint arXiv:1708.02002
   (2017)

\bibitem{luiten2021hota}
Luiten, J., Osep, A., Dendorfer, P., Torr, P., Geiger, A., Leal-Taix{\'e}, L.,
  Leibe, B.: Hota: A higher order metric for evaluating multi-object tracking.
  International journal of computer vision  \textbf{129},  548--578 (2021)

\bibitem{xiao2024motiontrack}
Xiao, C., Cao, Q., Zhong, Y., Lan, L., Zhang, X., Luo, Z., Tao, D.:
  Motiontrack: Learning motion predictor for multiple object tracking. Neural
  Networks  \textbf{179},  106539 (2024)

\bibitem{yi2024ucmctrack}
Yi, K., Luo, K., Luo, X., Huang, J., Wu, H., Hu, R., Hao, W.: Ucmctrack:
  Multi-object tracking with uniform camera motion compensation. In:
  Proceedings of the AAAI Conference on Artificial Intelligence. vol.~38, pp.
  6702--6710 (2024)

\bibitem{you2023utm}
You, S., Yao, H., Bao, B.K., Xu, C.: Utm: A unified multiple object tracking
  model with identity-aware feature enhancement. In: Proceedings of the
  IEEE/CVF Conference on Computer Vision and Pattern Recognition. pp.
  21876--21886 (2023)

\bibitem{zhang2022bytetrack}
Zhang, Y., Sun, P., Jiang, Y., Yu, D., Weng, F., Yuan, Z., Luo, P., Liu, W.,
  Wang, X.: Bytetrack: Multi-object tracking by associating every detection
  box. In: European conference on computer vision. pp. 1--21. Springer (2022)

\bibitem{zhang2021fairmot}
Zhang, Y., Wang, C., Wang, X., Zeng, W., Liu, W.: Fairmot: On the fairness of
  detection and re-identification in multiple object tracking. International
  journal of computer vision  \textbf{129},  3069--3087 (2021)

\bibitem{zhao2023representation}
Zhao, H., Wang, D., Lu, H.: Representation learning for visual object tracking
  by masked appearance transfer. In: Proceedings of the IEEE/CVF Conference on
  Computer Vision and Pattern Recognition. pp. 18696--18705 (2023)

\end{thebibliography}
\bibliographystyle{splncs04}

\end{document}